\documentclass[letterpaper, 10 pt, conference]{ieeeconf}
\IEEEoverridecommandlockouts
\usepackage{graphicx} 
\usepackage{subcaption} 
\usepackage[dvipsnames*,svgnames*,transparent]{xcolor}
\usepackage{amsmath} 

\usepackage{colortbl}
\usepackage{booktabs}
\usepackage{tabularx}
\usepackage{tabularray}
\usepackage{adjustbox}
\usepackage{multirow}
\usepackage{siunitx}
\usepackage{amsfonts}

\makeatletter
\let\NAT@parse\undefined
\makeatother
\usepackage[hidelinks]{hyperref} 
\usepackage{url}

\usepackage{tikz}
\usetikzlibrary{shadows}
\usetikzlibrary{shadows.blur}
\usetikzlibrary{arrows.meta}
\usepackage[skins]{tcolorbox}

\begin{document}
\title{\LARGE \bf RMP: A Random Mask Pretrain Framework for Motion Prediction}
\author{Yi~Yang$^{1,2}$, Qingwen~Zhang$^{1}$, Thomas~Gilles$^{3}$, Nazre~Batool$^{2}$, John~Folkesson$^{1}$
\thanks{$^{1}$Authors are with the Division of Robotics, Perception, and Learning (RPL), KTH Royal Institute of Technology, Stockholm 114 28, Sweden. (email: yiya@kth.se). 
}
\thanks{$^{2}$Authors are with Research and Development, Scania CV AB, Södertälje 151 87, Sweden.}
\thanks{$^{3}$Authors are with Mines Paris - PSL, Paris 75015, France}
}

\maketitle
Yi
\begin{abstract}
As the pretraining technique is growing in popularity, little work has been done on pretrained learning-based motion prediction methods in autonomous driving. 
In this paper, we propose a framework to formalize the pretraining task for trajectory prediction of traffic participants. 
Within our framework, inspired by the random masked model in natural language processing (NLP) and computer vision  (CV), objects’ positions at random timesteps are masked and then filled in by the learned neural network (NN).
By changing the mask profile, our framework can easily switch among a range of motion-related tasks.
We show that our proposed pretraining framework is able to deal with noisy inputs and improves the motion prediction accuracy and miss rate, especially for objects occluded over time by evaluating it on Argoverse and NuScenes datasets.
\end{abstract}

\section{Introduction}
Accurately predicting the motion of road users is essential in autonomous driving systems.
This predictive capability provides the planner with a forward-looking perspective on potential movements, thereby enhancing safety measures. 
While learning-based motion prediction has become increasingly popular in recent research, the exploration of pretraining and self-supervised learning within this field remains relatively limited.

The technique of random masking has demonstrated its effectiveness in various fields, such as natural language processing (NLP) and computer vision (CV), as evidenced by models like BERT \cite{bert} and Masked Autoencoders \cite{kaiming-mask-autoencoder} in conjunction with Vision Transformers (ViT \cite{vit}). 
Random masking involves concealing a portion of the data (masking), and then tasking the neural network with predicting the hidden elements, thereby creating a nontrivial and beneficial self-supervisory task. 
This method employs an asymmetric encoder-decoder architecture, which has proven to be particularly powerful regarding training speed with large datasets. Furthermore, it has demonstrated exceptional performance in transfer learning, particularly in tasks related to image processing.

\begin{figure}
    \centering
    \resizebox{0.45\textwidth}{!}{
        \begin{tikzpicture}[every shadow/.style={shadow xshift=3pt, shadow yshift=3pt,fill=gray!150!black}]
            \newcommand{\NROWS}{5}
            \newcommand{\NCOLS}{5}
            \newcommand{\NFUT}{2}
            \definecolor{KNOWNCOLOR}{RGB}{231,230,230}
            \definecolor{MASKCOLOR}{RGB}{181,198,229}
            \definecolor{GRIDCOLOR}{RGB}{175,171,171}
            \newcommand{\TEXTSCALE}{2}

            \begin{scope}[xshift=0cm,yshift=0cm]
                \fill [KNOWNCOLOR, drop shadow] (0,0) rectangle (\NCOLS,\NROWS);

                \fill [MASKCOLOR] (\NCOLS-\NFUT,0) rectangle (\NCOLS,\NROWS);

                \draw[->, >=latex, thick] (0,\NROWS+0.5) -- (\NCOLS,\NROWS+0.5) node[midway,above,scale=\TEXTSCALE] {Time};
                \draw[->, >=latex, thick] (-0.5,\NROWS) -- (-0.5,0) node[midway,left,scale=\TEXTSCALE] {\rotatebox{270}{Agents}};
                \draw[->, >=latex, thin] (-2,\NROWS-0.5) -- (-1,\NROWS-0.5) node[midway,above,scale=\TEXTSCALE] {ego};

                \draw [step=1cm, GRIDCOLOR, line width=3pt] (0,0) grid (\NCOLS,\NROWS);
            \end{scope}

            \begin{scope}[xshift=\NCOLS cm+3cm,yshift=0cm]
                \fill [KNOWNCOLOR, drop shadow] (0,0) rectangle (\NCOLS,\NROWS);

                \fill [MASKCOLOR] (2,0) rectangle (3,1);
                \fill [MASKCOLOR] (4,0) rectangle (5,1);
                \fill [MASKCOLOR] (0,1) rectangle (1,2);
                \fill [MASKCOLOR] (1,1) rectangle (2,2);
                \fill [MASKCOLOR] (3,2) rectangle (4,3);
                \fill [MASKCOLOR] (4,2) rectangle (5,3);
                \fill [MASKCOLOR] (1,4) rectangle (2,5);
                \fill [MASKCOLOR] (4,4) rectangle (5,5);

                \draw[->, >=latex, thick] (0,\NROWS+0.5) -- (\NCOLS,\NROWS+0.5) node[midway,above,scale=\TEXTSCALE] {Time};
                \draw[->, >=latex, thick] (-0.5,\NROWS) -- (-0.5,0) node[midway,left,scale=\TEXTSCALE] {\rotatebox{270}{Agents}};
                \draw[->, >=latex, thin] (-2,\NROWS-0.5) -- (-1,\NROWS-0.5) node[midway,above,scale=\TEXTSCALE] {ego};

                \draw [step=1cm, GRIDCOLOR, line width=3pt] (0,0) grid (\NCOLS,\NROWS);
            \end{scope}
        \end{tikzpicture}
        }
\caption{\textbf{Random Masking for Motion Data}. We treat time-sequential data as one dimension and all agents in the scenario as another, with each cell representing the high-dimensional features of an agent (including position, heading, agent type, agent shape, etc.). \textit{Left}: Motion prediction is a special case where all future timesteps are masked (shown in blue) \cite{sceneTransformer}. \textit{Right}: We apply random masking to a scenario, hiding patches for random agents and random time steps for pretraining. \textit{Ego} stands for the ego autonomous vehicle.}
\label{fig:idea}
\end{figure}
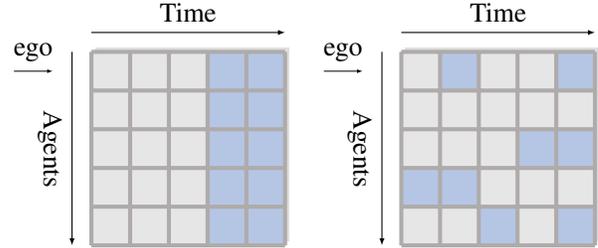

Inspired by SceneTransformer \cite{sceneTransformer}, the motion prediction task is linked with a mask on the future time sequential data of road users. 
As depicted in Fig. \ref{fig:idea}, the data for all agents can be represented as a grid, with time and agent forming the two axes. In this context, motion prediction becomes a unique task wherein future states are masked \cite{sceneTransformer}. 
This leads us to the natural question: \textit{Could random mask pretraining be effectively applied to general motion tasks as well?} 
These tasks include motion prediction (marginal, conditional, etc.), occlusion handling, and others. We introduce a straightforward yet potent framework for random masking pretraining (RMP) for motion tasks. 
Our RMP selectively conceals motion patches, allowing the random mask to capture spatial and social correlations among all agents in a given scenario. This universal framework can be readily integrated into numerous motion prediction methodologies. In this paper, we demonstrate its adaptability by incorporating it into several state-of-the-art models, including Autobots \cite{girgis2022latent-autobot} and Hivt \cite{zhou2022hivt}.

We assess the impact of pretraining on performing three different tasks: motion prediction, conditional motion prediction, and occlusion handling. 
In case of conditional motion prediction, not only is the historical information of all agents provided, but also the desired trajectory of the ego vehicle. The network then endeavors to predict the trajectories of all other agents.

In addition to classic motion prediction, we also treat  occlusion handling as a separate task to evaluate our proposed framework. In real-world scenarios, occlusions are a common occurrence where one or more agents are partially or entirely obscured from view. 
Under such circumstances, predicting the motion of the occluded 
agents become a complex task that can significantly influence the overall performance of the autonomous driving system, especially with occlusions happening over short distances. This is a nontrivial issue that has often not been specifically focused on in practice. For agents whose historical trajectories are partially or heavily occluded, we evaluate the performance of the current state-of-the-art networks with and without masking pretraining in an object-based manner.

Our experimental results indicate that motion prediction benefits from transfer learning for generalization and random masking. Our framework demonstrates effective performance on the Argoverse \cite{Argoverse} and NuScenes \cite{caesar2020nuscenes} datasets. Our code will be publicly accessible at \textcolor{blue}{\url{https://github.com/KTH-RPL/RMP}}.

In this paper, we make the following contributions:
\begin{itemize}
\item We introduce a pretraining framework for a range of motion-related tasks.
\item We design experiments to validate the effectiveness of random masking.
\item We highlight that occlusion handling remains a challenge for current state-of-the-art methods and demonstrate that our pretraining method enhances performance in this area.
\end{itemize}

\section{Related Work}
\subsection{Motion Prediction}
Motion prediction has recently been explored rapidly with large open datasets and public benchmarks \cite{Argoverse,caesar2020nuscenes,ettinger2021large_waymo,zhan2019interaction}. Early approaches drew inspiration from successful computer vision techniques, where map and agents' historical trajectories were rasterized into images using specific color encoding \cite{bansal2018chauffeurnet, chai2019multipath, Hong_2019_CVPR}.
However, rasterization carries certain limitations, such as the challenge of selecting an optimal Field-Of-View (FOV) due to the high computational cost of high-resolution imaging and the potential for long-distance information loss. An alternative approach to these challenges is using sparse vectors and polygons, as exemplified by VectorNet \cite{gao2020vectornet}.
Other network architectures that have been explored include Graph Neural Networks \cite{liang2020lanegcn, huang2019stgat} and Transformers \cite{sceneTransformer, liu2021mmtransformer, zhou2022hivt, huang2022multi}. The outputs of these representations vary: some generate a set of point trajectories in an end-to-end manner \cite{liang2020lanegcn, sceneTransformer, zhou2022hivt}, while others generate top K trajectory samples from anchors \cite{chai2019multipath}, heatmaps \cite{gilles2021home, gilles2022gohome, gilles2021thomas}, or kinematic models \cite{ma2019wasserstein, varadarajan2022multipath++}. 
Owing to its adaptability, our proposed framework can be effectively incorporated into many of these methods.

\subsection{Self-supervised Learning}
Self-supervised learning methods have garnered substantial interest across various fields, such as NLP and CV \cite{bert, caron2020unsupervised, grill2020bootstrap, bao2021beit}. These methods leverage different tasks to initialize network weights in the pretraining phase. For instance, contrastive learning \cite{hjelm2018learning, oord2018representation} designs tasks that distinguish between similarities and dissimilarities, utilizing both original data samples and their augmented counterparts.
The Masked Autoencoder, proposed by \cite{kaiming-mask-autoencoder}, uses a masking encoder to reconstruct missing pixels in images during the pretraining phase, resulting in better performance and a training speed that is four times faster than training from scratch. This technique has inspired applications in a variety of domains, such as video \cite{tong2022videomae, feichtenhofer2022masked}, 3D point clouds \cite{yu2022point}, and visual reinforcement learning in robotics \cite{pmlr-v205-seo23a}.
Self-supervised learning for motion prediction in autonomous driving remains largely unexplored. 
However, in the past year, a few studies have started investigating this area \cite{xu2022pretram, wagnerroad, azevedo2022exploiting, bhattacharyya2022ssllanes}.
Prarthana et al.~\cite{bhattacharyya2022ssllanes} propose a suite of four pretraining tasks, including lane masking, intersection distance calculation, maneuver classification, and success/failure classification.
The work most similar to ours is the recent archive preprint \cite{chen2023trajmae} which shows results similar to our own on one of the tasks we tested (prediction).
Our work here was developed independently to \cite{chen2023trajmae}.

\subsection{Conditional Motion Prediction}
Compared to standard motion prediction, conditional motion prediction offers additional information by incorporating specific conditions, such as the intended path of the ego vehicle.
For example, the work presented in \cite{khandelwal2020if} generates predictions based on hypothetical `what-if' interactions among road users and lanes. In this way, although their targeted task closely resembles standard motion prediction, it extends the context by incorporating speculative interaction scenarios.
Additionally, studies like \cite{gu2021densetnt} and \cite{gilles2021thomas} adopt a two-step approach in their prediction methodology by first predicting the destination positions, which are then used as conditions for predicting full trajectories. This effectively transforms the prediction task into a conditional one, where the trajectories are predicated on hypothesized destinations.

\subsection{Occlusion Handling} 
Handling occlusions in motion prediction is crucial for enhancing the robustness and reliability of autonomous driving systems.
A widely adopted representation called Occupancy Grid Map (OGM) captures the spatial arrangement of obstacles and free space where each grid cell represents the estimated probability of an agent's presence within. Predicting future OGM allows the formation of occluded areas, thus offering a more comprehensive understanding of the environment \cite{itkina2022multi, lange2022lopr}. Nevertheless, these approaches based on OGM can be computationally expensive, particularly for high-resolution, large, and complex environments.
For object-based methods, there has been limited work due to the lack of motion prediction datasets that annotate occluded objects. Most datasets are primarily collected from the ego vehicle's perspective \cite{Argoverse, caesar2020nuscenes}.
To help mitigate this, we have post-processed the INTERACTION dataset \cite{zhan2019interaction}, which was captured from bird's-eye-view drones. This has allowed us to estimate occlusion labels for objects, and we openly share the resulting post-processed dataset for further research in this area.

\section{Problem Formulation}
\label{AA}
Consider a scenario including $N$ agents' trajectories $A$ over $T$ timestamps, denoted as $A_i \in \mathbb{R} ^{T\times D_{agent}}$, where $i \in [1, N]$, along with the surrounding road topology $Map \in \mathbb{R} ^{S \times P \times D_{road}}$. Here, $S$ represents the number of road segments, $P$ denotes the number of points within a segment, and $D$ signifies the vector feature dimension that includes position coordinates $x,y$ and the validity mask for both $D_{agent}$ and $D_{road}$. If yaw angle, velocity and agent size of the agents are provided in the dataset, they are also added into the feature $D$.

In the context of motion prediction, we are provided with the historical trajectory $A_{history} \in \mathbb{R} ^{T_{obs}\times D_{agent}}$, where $T_{obs}$ signifies the observed historical timestamps, and our task is to predict the future trajectory $A_{future} \in \mathbb{R} ^{T_{fut}\times D_{agent}}$.

Here, it is worth mentioning that occlusion can complicate this task, as $A_{history}$ may contain many occluded objects with unknown states. In the case of conditional motion prediction, however, additional elements are taken into account. In particular, the historical information is supplemented with the ego vehicle's anticipated future route path $A_{ego} \in \mathbb{R} ^{T_{fut}\times D_{agent}}$ (where $i$ equals to index of ego vehicle), which forms part of the input.

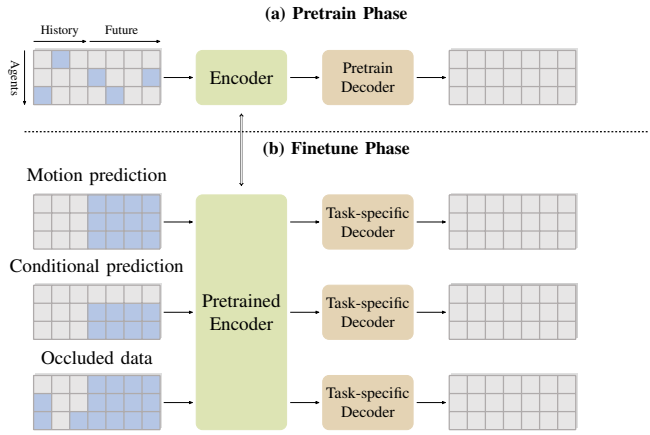
\begin{figure}
    \centering
    \resizebox{\linewidth}{!}{
        \begin{tikzpicture}[every shadow/.style={shadow xshift=3pt, shadow yshift=3pt,fill=gray!150!black}]
            \newcommand{\NROWS}{3}
            \newcommand{\NCOLS}{7}
            \newcommand{\NPAST}{3}
            \definecolor{MASKCOLOR}{RGB}{181,198,229}
            \definecolor{KNOWNCOLOR}{RGB}{231,230,230}
            \definecolor{GRIDCOLOR}{RGB}{175,171,171}

            \newcommand{\GRIDWIDTH}{2pt}
            
            \newcommand{\GAP}{2}
            \newcommand{\VGAP}{2}
            \definecolor{ENCCOLOR}{RGB}{221,228,180}
            \definecolor{DECCOLOR}{RGB}{228,211,180}
            \newcommand{\MODWIDTH}{5}
            \newcommand{\TEXTSCALE}{3}
            \newcommand{\PHASETEXTSCALE}{2.8}
            \newcommand{\IMGWIDTH}{\NCOLS*3/2}

            \begin{scope}[xshift=0cm,yshift=0cm]
                \fill [KNOWNCOLOR, drop shadow] (0,0) rectangle (\NCOLS,\NROWS);
                \node[text=black,scale=\PHASETEXTSCALE,font=\bfseries] at (\NCOLS/2+\GAP*3/2+\MODWIDTH+\IMGWIDTH/2, \NROWS+\VGAP) {(a) Pretrain Phase};
                \node[text=black,scale=\PHASETEXTSCALE,font=\bfseries] at (\NCOLS/2+\GAP*3/2+\MODWIDTH+\IMGWIDTH/2, -\VGAP*3/2+\VGAP*1/4) {(b) Finetune Phase};

                \draw[-{Latex[round]}, >=latex, thick] (0,\NROWS+0.5) -- (\NPAST-0.1,\NROWS+0.5) node[midway,above,scale=2] {History};
                \draw[-{Latex[round]}, >=latex, thick] (\NPAST+0.1,\NROWS+0.5) -- (\NCOLS,\NROWS+0.5) node[midway,above,scale=2] {Future{\color{white}y}};
                \draw[-{Latex[round]}, >=latex, thick] (-0.5,\NROWS) -- (-0.5,0) node[midway,left,scale=2] {\rotatebox{270}{Agents}};
                
                \fill [color=MASKCOLOR] (0,0) rectangle (1,1);
                \fill [MASKCOLOR] (4,0) rectangle (5,1);
                \fill [MASKCOLOR] (3,1) rectangle (4,2);
                \fill [MASKCOLOR] (6,1) rectangle (7,2);
                \fill [MASKCOLOR] (1,2) rectangle (2,3);

                \draw [step=1cm, GRIDCOLOR, line width=\GRIDWIDTH] (0,0) grid (\NCOLS,\NROWS);

                \draw[ENCCOLOR, rounded corners=10pt, fill=ENCCOLOR] (\NCOLS+\GAP,0) rectangle (\NCOLS+\GAP+\MODWIDTH,\NROWS);
                \node[text=black,scale=\TEXTSCALE] at (\NCOLS+\GAP+\MODWIDTH/2,\NROWS/2) {Encoder};

                \draw[DECCOLOR, rounded corners=10pt, fill=DECCOLOR] (\NCOLS+\GAP+\GAP+\MODWIDTH,0) rectangle (\NCOLS+\GAP+\GAP+\MODWIDTH+\MODWIDTH,\NROWS);
                \node[text=black,scale=2.5, align=center] at (\NCOLS+2*\GAP+\MODWIDTH*3/2,\NROWS/2) {Pretrain \\ Decoder};

                \fill [KNOWNCOLOR, drop shadow] (\NCOLS+\GAP+\GAP+\GAP+\MODWIDTH+\MODWIDTH,0) rectangle (\NCOLS+\GAP+\GAP+\GAP+\MODWIDTH+\MODWIDTH+\NCOLS,\NROWS);
                \draw [step=1cm, GRIDCOLOR, line width=\GRIDWIDTH] (\NCOLS+\GAP+\GAP+\GAP+\MODWIDTH+\MODWIDTH,0) grid (\NCOLS+\GAP+\GAP+\GAP+\MODWIDTH+\MODWIDTH+\NCOLS,\NROWS);

                \draw[-{Latex[round]}, thick] (\NCOLS+0.2,\NROWS/2) -- (\NCOLS+\GAP-0.2,\NROWS/2);
                \draw[-{Latex[round]}, thick] (\NCOLS+\GAP+\MODWIDTH+0.2,\NROWS/2) -- (\NCOLS+2*\GAP+\MODWIDTH-0.2,\NROWS/2);
                \draw[-{Latex[round]}, thick] (\NCOLS+2*\GAP+2*\MODWIDTH+0.2,\NROWS/2) -- (\NCOLS+3*\GAP+2*\MODWIDTH-0.2,\NROWS/2);

                \draw[dashed, line width=2pt] (-0.5,-\VGAP+\VGAP*1/4) -- (\NCOLS+\IMGWIDTH+3*\GAP+2*\MODWIDTH+0.5,-\VGAP+\VGAP*1/4);

                \draw[Implies-Implies,double equal sign distance,thick] (\NCOLS+\GAP+\MODWIDTH/2,-\VGAP/4) to (\NCOLS+\GAP+\MODWIDTH/2,-\VGAP*9/4);

            \end{scope}

            \begin{scope}[xshift=0cm,yshift=-\NROWS cm-2.5*\VGAP cm]
                \newcommand{\NFUT}{4}
                \node[text=black,scale=\TEXTSCALE] at (\NCOLS/2, \NROWS+\VGAP/2) {Motion prediction};
            
                \fill [KNOWNCOLOR, drop shadow] (0,0) rectangle (\NCOLS,\NROWS);

                \fill [MASKCOLOR] (\NCOLS-\NFUT,0) rectangle (\NCOLS,\NROWS);

                \draw [step=1cm, GRIDCOLOR, line width=\GRIDWIDTH] (0,0) grid (\NCOLS,\NROWS);

                \draw[ENCCOLOR, rounded corners=10pt, fill=ENCCOLOR] (\NCOLS+\GAP,-2*\NROWS-2*\VGAP) rectangle (\NCOLS+\GAP+\MODWIDTH,\NROWS);
                \node[text=black,scale=\TEXTSCALE, align=center] at (\NCOLS+\GAP+\MODWIDTH/2,-\NROWS/2-2*\VGAP/2) {Pretrained \\ Encoder};

                \draw[DECCOLOR, rounded corners=10pt, fill=DECCOLOR] (\NCOLS+\GAP+\GAP+\MODWIDTH,0) rectangle (\NCOLS+\GAP+\GAP+\MODWIDTH+\MODWIDTH,\NROWS);
                \node[text=black,scale=2.5, align=center] at (\NCOLS+2*\GAP+\MODWIDTH*3/2,\NROWS/2) {Task-specific \\ Decoder};

                \fill [KNOWNCOLOR, drop shadow] (\NCOLS+\GAP+\GAP+\GAP+\MODWIDTH+\MODWIDTH,0) rectangle (\NCOLS+\GAP+\GAP+\GAP+\MODWIDTH+\MODWIDTH+\NCOLS,\NROWS);
                \draw [step=1cm, GRIDCOLOR, line width=\GRIDWIDTH] (\NCOLS+\GAP+\GAP+\GAP+\MODWIDTH+\MODWIDTH,0) grid (\NCOLS+\GAP+\GAP+\GAP+\MODWIDTH+\MODWIDTH+\NCOLS,\NROWS);

                \draw[-{Latex[round]}, thick] (\NCOLS+0.2,\NROWS/2) -- (\NCOLS+\GAP-0.2,\NROWS/2);
                \draw[-{Latex[round]}, thick] (\NCOLS+\GAP+\MODWIDTH+0.2,\NROWS/2) -- (\NCOLS+2*\GAP+\MODWIDTH-0.2,\NROWS/2);
                \draw[-{Latex[round]}, thick] (\NCOLS+2*\GAP+2*\MODWIDTH+0.2,\NROWS/2) -- (\NCOLS+3*\GAP+2*\MODWIDTH-0.2,\NROWS/2);
                
            \end{scope}

            \begin{scope}[xshift=0cm,yshift=-2*\NROWS cm-3.5*\VGAP cm]
                \newcommand{\NFUT}{4}

                \node[text=black,scale=\TEXTSCALE] at (\NCOLS/2, \NROWS+\VGAP/2) {Conditional prediction};
            
                \fill [KNOWNCOLOR, drop shadow] (0,0) rectangle (\NCOLS,\NROWS);

                \fill [MASKCOLOR] (\NCOLS-\NFUT,0) rectangle (\NCOLS,\NROWS);
                \fill [KNOWNCOLOR] (\NCOLS-\NFUT,\NROWS-1) rectangle (\NCOLS,\NROWS);

                \draw [step=1cm, GRIDCOLOR, line width=\GRIDWIDTH] (0,0) grid (\NCOLS,\NROWS);

                \draw[DECCOLOR, rounded corners=10pt, fill=DECCOLOR] (\NCOLS+\GAP+\GAP+\MODWIDTH,0) rectangle (\NCOLS+\GAP+\GAP+\MODWIDTH+\MODWIDTH,\NROWS);
                \node[text=black,scale=2.5, align=center] at (\NCOLS+2*\GAP+\MODWIDTH*3/2,\NROWS/2) {Task-specific \\ Decoder};

                \fill [KNOWNCOLOR, drop shadow] (\NCOLS+\GAP+\GAP+\GAP+\MODWIDTH+\MODWIDTH,0) rectangle (\NCOLS+\GAP+\GAP+\GAP+\MODWIDTH+\MODWIDTH+\NCOLS,\NROWS);

                \fill [KNOWNCOLOR, drop shadow] (\NCOLS+\GAP+\GAP+\GAP+\MODWIDTH+\MODWIDTH,0) rectangle (\NCOLS+\GAP+\GAP+\GAP+\MODWIDTH+\MODWIDTH+\NCOLS,\NROWS);
                \draw [step=1cm, GRIDCOLOR, line width=\GRIDWIDTH] (\NCOLS+\GAP+\GAP+\GAP+\MODWIDTH+\MODWIDTH,0) grid (\NCOLS+\GAP+\GAP+\GAP+\MODWIDTH+\MODWIDTH+\NCOLS,\NROWS);
                \draw[-{Latex[round]}, thick] (\NCOLS+0.2,\NROWS/2) -- (\NCOLS+\GAP-0.2,\NROWS/2);
                \draw[-{Latex[round]}, thick] (\NCOLS+\GAP+\MODWIDTH+0.2,\NROWS/2) -- (\NCOLS+2*\GAP+\MODWIDTH-0.2,\NROWS/2);
                \draw[-{Latex[round]}, thick] (\NCOLS+2*\GAP+2*\MODWIDTH+0.2,\NROWS/2) -- (\NCOLS+3*\GAP+2*\MODWIDTH-0.2,\NROWS/2);
                
            \end{scope}

            \begin{scope}[xshift=0cm,yshift=-3*\NROWS cm-4.5*\VGAP cm]
                \newcommand{\NFUT}{4}

                \node[text=black,scale=\TEXTSCALE] at (\NCOLS/2, \NROWS+\VGAP/2) {Occluded data};
            
                \fill [KNOWNCOLOR, drop shadow] (0,0) rectangle (\NCOLS,\NROWS);

                \fill [MASKCOLOR] (\NCOLS-\NFUT,0) rectangle (\NCOLS,\NROWS);
                \fill [MASKCOLOR] (0,0) rectangle (1,1);
                \fill [MASKCOLOR] (2,0) rectangle (3,1);
                \fill [MASKCOLOR] (0,1) rectangle (1,2);
                \fill [MASKCOLOR] (3,2) rectangle (4,3);

                \draw [step=1cm, GRIDCOLOR, line width=\GRIDWIDTH] (0,0) grid (\NCOLS,\NROWS);

                \draw[DECCOLOR, rounded corners=10pt, fill=DECCOLOR] (\NCOLS+\GAP+\GAP+\MODWIDTH,0) rectangle (\NCOLS+\GAP+\GAP+\MODWIDTH+\MODWIDTH,\NROWS);
                \node[text=black,scale=2.5, align=center] at (\NCOLS+2*\GAP+\MODWIDTH*3/2,\NROWS/2) {Task-specific \\ Decoder};

                \fill [KNOWNCOLOR, drop shadow] (\NCOLS+\GAP+\GAP+\GAP+\MODWIDTH+\MODWIDTH,0) rectangle (\NCOLS+\GAP+\GAP+\GAP+\MODWIDTH+\MODWIDTH+\NCOLS,\NROWS);
                \draw [step=1cm, GRIDCOLOR, line width=\GRIDWIDTH] (\NCOLS+\GAP+\GAP+\GAP+\MODWIDTH+\MODWIDTH,0) grid (\NCOLS+\GAP+\GAP+\GAP+\MODWIDTH+\MODWIDTH+\NCOLS,\NROWS);

                \draw[-{Latex[round]}, thick] (\NCOLS+0.2,\NROWS/2) -- (\NCOLS+\GAP-0.2,\NROWS/2);
                \draw[-{Latex[round]}, thick] (\NCOLS+\GAP+\MODWIDTH+0.2,\NROWS/2) -- (\NCOLS+2*\GAP+\MODWIDTH-0.2,\NROWS/2);
                \draw[-{Latex[round]}, thick] (\NCOLS+2*\GAP+2*\MODWIDTH+0.2,\NROWS/2) -- (\NCOLS+3*\GAP+2*\MODWIDTH-0.2,\NROWS/2);
                
            \end{scope}
        \end{tikzpicture}
    }
    \caption{The pretraining framework. 
    In the first pretrain phase, all agents' information including the history and future time are concatenated together. Next, random masking is applied. Then, given incomplete information about agents' positions with time (in grey), where some positions are randomly masked (in blue), the network trains to fill in the missing positions. In the fine-tuning phase, there are three tasks that correspond to three special masking cases. Once trained, the pretrained encoder is used for different tasks.
    }
    \label{fig:network}
\end{figure}

\section{Methodology}
In this section, we outline the strategy employed in our study. Fig. \ref{fig:network} provides an illustration of the complete training framework, and the specifics of the random masking application are outlined in the following sub-sections.

\begin{figure}[t]
    \centering
    \begin{subfigure}[b]{0.45\textwidth}
        \centering
        \includegraphics[width=\textwidth]{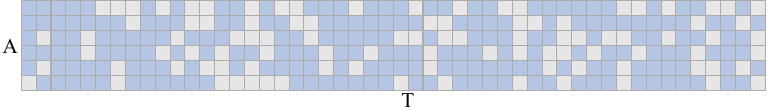}
        \vspace{-13pt}
        \caption{Pointwise, 75\%}
        \label{fig:mask_pointwise}
    \end{subfigure}
    \begin{subfigure}[b]{0.45\textwidth}
        \vspace{1em}
        \centering
        \includegraphics[width=\textwidth]{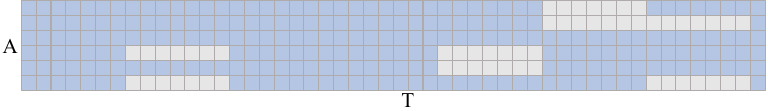}
        \vspace{-13pt}
        \caption{Patchwise, 75\%}
        \label{fig:mask_patchwise}
    \end{subfigure}
    \begin{subfigure}[b]{0.45\textwidth}
        \vspace{1em}
        \centering
        \includegraphics[width=\textwidth]{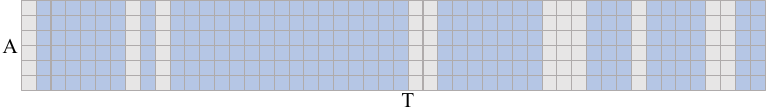}
        \vspace{-13pt}
        \caption{Time-based, 75\%}
        \label{fig:mask_time_only}
    \end{subfigure}
    \caption{Different mask sampling strategies: (a) random pointwise masking, (b) random patchwise masking for random agents, (c) random masking in time. All show 75\% masking in total (in blue)  and the remaining data (in grey) will be fed into the network.}
    \label{fig:mask_strategy}
\end{figure}

\renewcommand{\arraystretch}{1.3}
\begin{table*}
\centering
\caption{Ablation experiments on our pretrain framework with the Autobot model on Argoverse validation dataset. We have evaluated the influences of different mask sampling strategies, finetuning with or without the frozen pretrained encoder weights, and also the encoder size. \textit{w/[P]} represents the method with our random masking pretraining. The default setting is highlighted in grey.}
\begin{subtable}{\linewidth}
\centering
\caption{Different mask ratios and profiles where 75\% pointwise mask performs best.}
\label{tab:mask_ratio}
\begin{tabular}{lccc|ccc} 
\hline
\multicolumn{1}{c}{\multirow{2}{*}{mask ratio}} & \multicolumn{3}{c|}{minADE\_6 ↓}                              & \multicolumn{3}{c}{minFDE\_6 ↓}                                \\ 
\cline{2-7}
\multicolumn{1}{c}{}                            & point                                     & patch & time-only & point                                     & patch & time-only  \\ 
\hline
Autobot w/ [P] 25 \%~                           & 0.725                                     & 0.720 & 0.722     & 1.308                                     & 1.302 & 1.297      \\
Autobot w/ [P] 50 \%~                           & 0.720                                     & 0.727 & 0.727     & \textbf{1.292}                                     & 1.306 & 1.312      \\
Autobot w/ [P] 75 \%                            & {\cellcolor[rgb]{0.906,0.902,0.902}}\textbf{0.717} & 0.728 & 0.726     & {\cellcolor[rgb]{0.906,0.902,0.902}}\textbf{1.292} & 1.318 & 1.312      \\
\hline \\[3pt] 
\end{tabular}
\end{subtable}

\begin{minipage}{\textwidth}
\begin{subtable}{0.4\linewidth}
\centering
\caption{With or without frozen pretrained encoder.}
\label{tab:frozen_encoder}
\begin{tabular}{lllll} 
\toprule
\multicolumn{1}{c}{\multirow{2}{*}{Argoverse}} & \multicolumn{2}{c}{minADE\_6 ↓}                             & \multicolumn{2}{c}{minFDE\_6 ↓}                              \\ 
\cline{2-5}
\multicolumn{1}{c}{}                           & Frozen & Unfrozen                                           & Frozen & Unfrozen                                            \\ 
\hline
Autobot                                        & -      & 0.766                                              & -      & 1.38                                                \\
Autobot w/ [P] 25 \%~                          & 0.797  & 0.725                                              & 1.517  & 1.308                                               \\
Autobot w/ [P] 50 \%~                          & 0.793  & 0.720                                              & 1.499  & \textbf{1.292}                                      \\
Autobot w/ [P] 75 \%                           & 0.793  & {\cellcolor[rgb]{0.906,0.902,0.902}}\textbf{0.717} & 1.504  & {\cellcolor[rgb]{0.906,0.902,0.902}}\textbf{1.292}  \\
\bottomrule
\end{tabular}
\end{subtable}
\hspace{2.5cm}
\begin{subtable}{0.4\linewidth}
\centering
\caption{Different number of encoder blocks.}
\label{tab:encoder_size}
\begin{tabular}{ccc} 
\toprule
blocks & minADE\_6 ↓                               & minFDE\_6 ↓                                \\ 
\hline
2      & {\cellcolor[rgb]{0.906,0.902,0.902}}\textbf{0.717} & {\cellcolor[rgb]{0.906,0.902,0.902}}\textbf{1.292}  \\
4      & 0.736                                     & 1.346                                      \\
6      & 0.727                                     & 1.314                                      \\
\bottomrule
\end{tabular}
\end{subtable}
\end{minipage}
\label{tab:ablation_pointwise_mask}
\end{table*}
\renewcommand{\arraystretch}{1.0}

\subsection{Network}
Our approach is an extension of the masked autoencoder \cite{bert, kaiming-mask-autoencoder} for time-sequential trajectory data and aims to provide a simple, yet effective framework that is applicable to many motion prediction methodologies with minimal domain-specific knowledge required.

The framework can accommodate many network architectures in a two-stage process.
In the first stage, different masking strategies are applied to all timestamps including the history and future timestamps, and for all agents. Given incomplete waypoints, the model tries to predict $K$ possible completed trajectories. Therefore, we don't need to change the loss function from the original methods. 
In the second fine-tuning stage,  the network combines the pretrained encoder and the task-specific decoder. 

Our method tests on two networks -Autobot-Joint \cite{girgis2022latent-autobot} and HiVT \cite{zhou2022hivt}. Autobot-Joint \cite{girgis2022latent-autobot} is a transformer-based network that uses an axial attention mechanism to learn the temporal and spatial correlations among agents and road topology. 
Hivt \cite{zhou2022hivt} models the local and global context in a translation and rotation invariant transformer network.

\subsection{Masking}
By changing the validity mask within the input, the pretraining task can easily be  switched among trajectory completion (pretraining task), motion prediction, and conditional prediction. The mask defines which parts can be seen by the network. For the unseen parts, we further set them as zeros to guarantee a clean input for the network.

The random masking pretraining incorporates pointwise, patchwise, and time-based strategies, as illustrated in Fig. \ref{fig:mask_strategy}, each serving a distinct purpose. 
The pointwise approach (Fig. \ref{fig:mask_pointwise}) primarily facilitates the learning of interpolation and correlation over a short period from noisy data. 
In contrast, the patchwise method (Fig. \ref{fig:mask_patchwise}) fosters an understanding of interactions over extended periods. Inspired by the masked autoencoder approach to video data \cite{feichtenhofer2022masked, tong2022videomae}, each agent's trajectory is divided into non-overlapping patches in space and time given a certain timeframe. The size of these patches is chosen randomly, and patches are masked randomly. 
The time-based strategy (Fig. \ref{fig:mask_time_only}) simulates scenarios where a sensor might fail abruptly, leading to missing data at random timestamps.

The three tasks - motion prediction, conditional prediction and occlusion handling are three special masking cases (Fig. \ref{fig:network}). Each task involves the process of prediction, where future trajectories are treated as unknown and masked out. In conditional motion prediction, alongside the full historical data, the future desired path of the ego vehicle is also provided. For occlusion handling, the input data is often incomplete due to occlusions. 
Since the three tasks correspond to special cases of masking, they can be carried out by adapting the same network architecture accordingly.

\begin{figure}
  \centering
  \begin{subfigure}[b]{0.22\textwidth}
    \includegraphics[width=\textwidth]{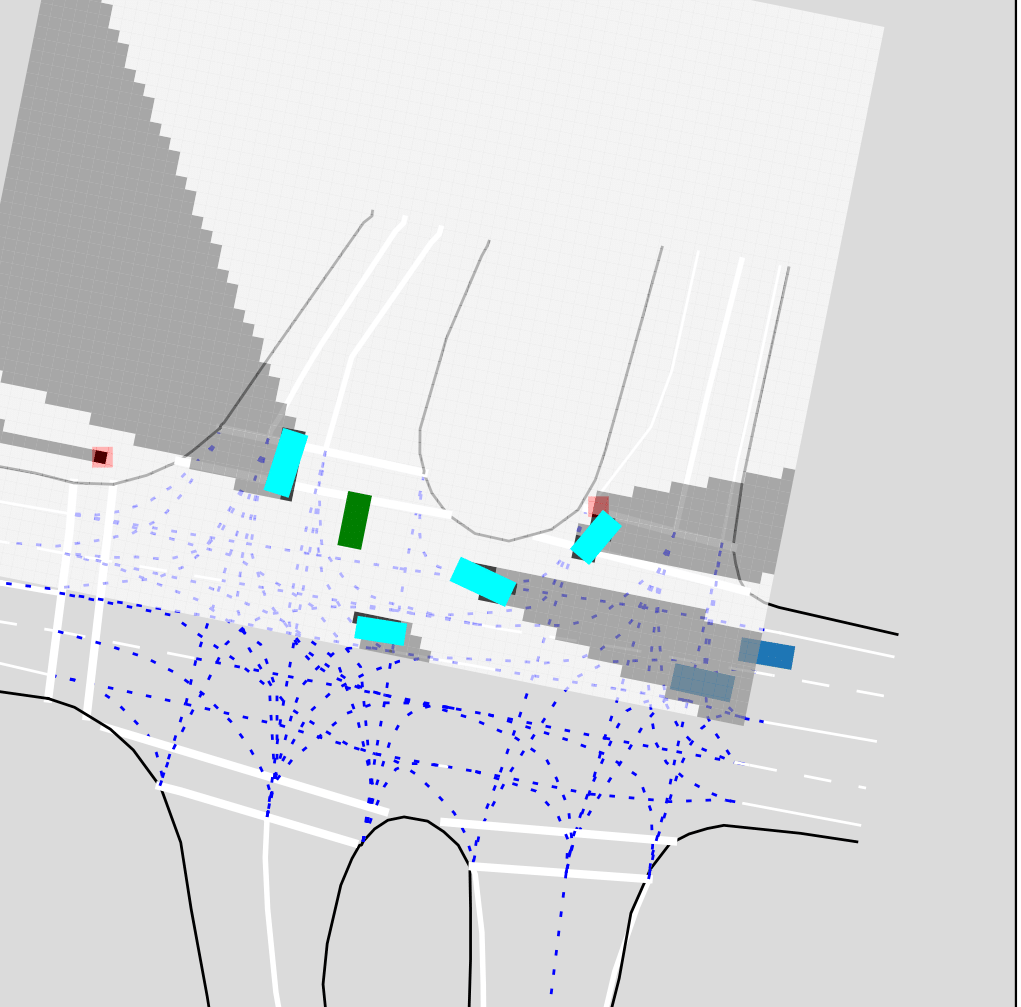}
  \end{subfigure}
  \begin{subfigure}[b]{0.23\textwidth}
    \includegraphics[width=\textwidth]{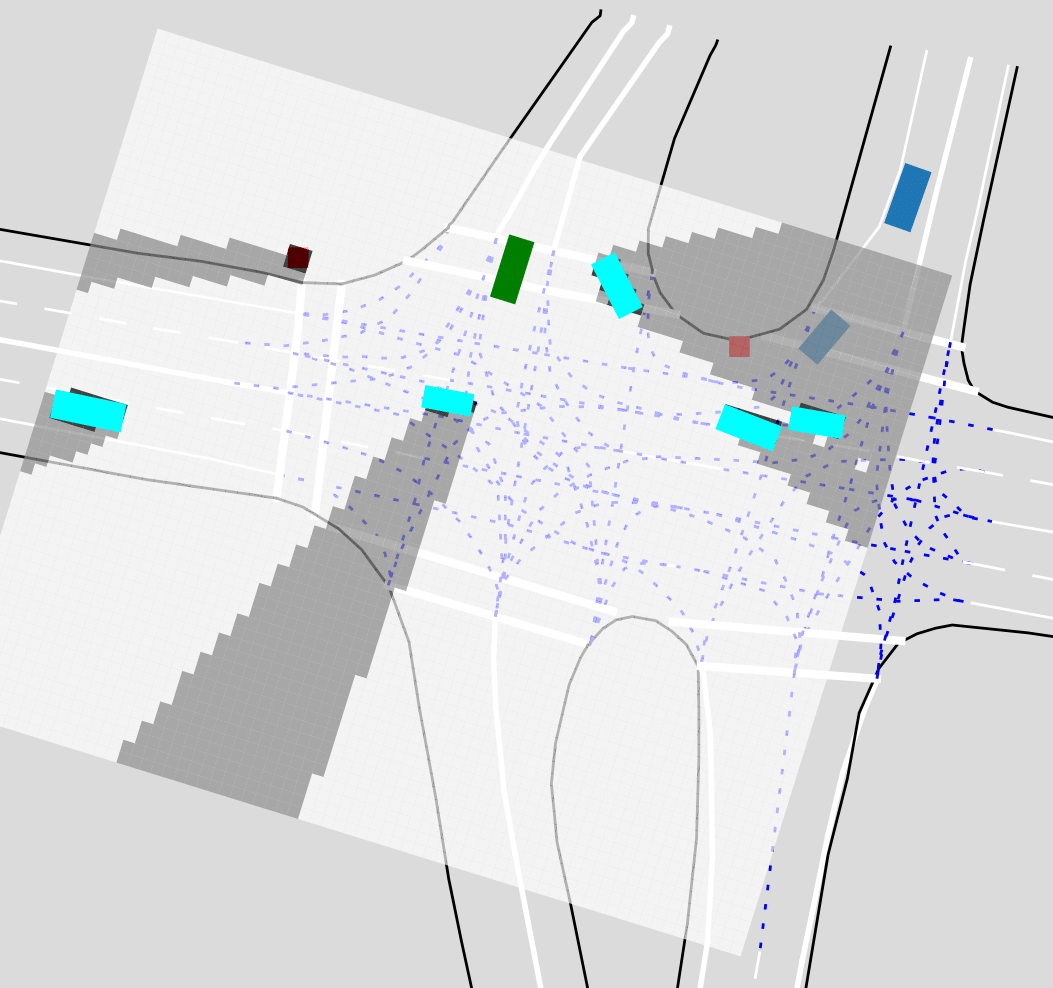}
  \end{subfigure}
  \caption{Two examples of labeling occluded objects using ray tracing occupancy grid map from one vehicle's view. The labeled object track will be used to evaluate the occlusion handling performance.
  The dark blue occluded agent in the occluded area (in \textcolor{gray}{grey} grids) is blocked by other visible agents (in \textcolor{cyan}{cyan}), from the ego vehicle's (in \textcolor{teal}{teal}) view. 
  }
  \label{fig:label_occlusion}
\end{figure}

\begin{figure*}[h]
    \centering
    \begin{subfigure}[b]{\textwidth}
        \includegraphics[width=\textwidth]{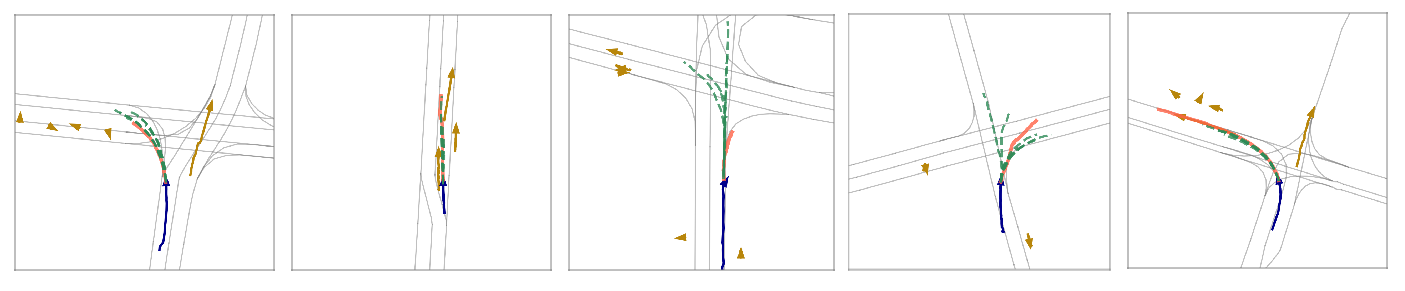}
        \caption{Qualitative results of motion prediction using Autobot with random mask pretrain on Argoverse dataset. The past trajectories of all other vehicles are shown in brown, the past trajectories of ego vehicle are shown in dark blue, the ground-truth trajectories are shown in red, the predicted trajectories are shown in green.}
    \end{subfigure}

    \begin{subfigure}[b]{\textwidth}
        \includegraphics[width=\textwidth]{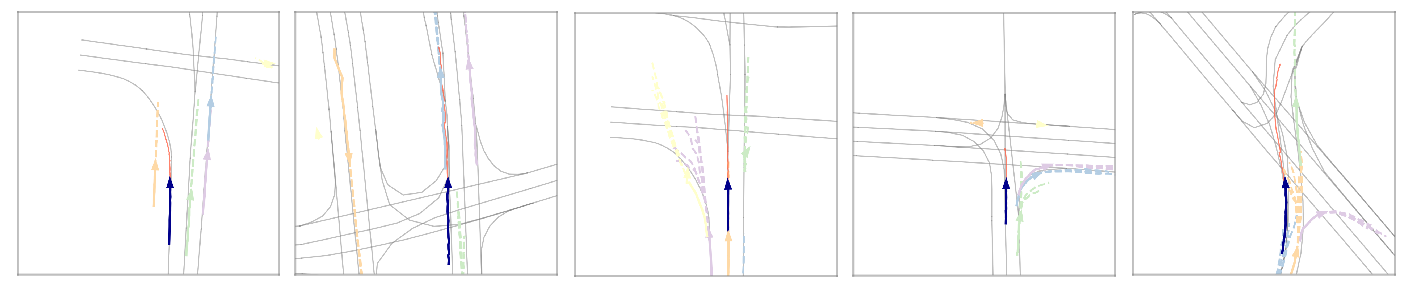}
        \caption{Qualitative results of conditional motion prediction using Autobot with random mask pretrain on Argoverse dataset. Given the ego vehicle's past (in dark blue) and future (in red) trajectories, and past trajectories of all other vehicles (solid line, one color for one agent), the predicted trajectories of all other agents are shown in the dashed line.}
    \end{subfigure}
    
    \caption{Qualitative results with our random mask pretrain framework.}
    \label{fig:qualitative_result}
\end{figure*}

\section{Experiments}
\subsection{Datasets}
\label{sec:exp-dataset}
We evaluate the efficacy of our pretraining framework for motion and conditional prediction on two widely used datasets: Argoverse \cite{Argoverse} and nuScenes \cite{caesar2020nuscenes}.
Argoverse motion forecasting dataset contains $205,942$ training sequences and $39,472$ validation sequences. 
Each sequence includes data of all agents' positions over a 5 seconds period at $10~\si{Hz}$. 
The task is to predict the subsequent 3 seconds' trajectory based on the initial 2 seconds of past observations with HD map information provided. 
The nuScenes dataset consists of $32,186$ training and $9,041$ validation sequences. 
The objective, in this case, is to predict future 6 seconds' trajectories at a rate of 2 Hz, given the past 2 seconds' trajectory data.

In order to evaluate our model's proficiency in handling occlusions, we leverage the multi-track INTERACTION dataset \cite{zhan2019interaction}. This dataset is collected by drones and potential traffic cameras, which enables the potential to label occluded objects from the perspective of a single vehicle.
We auto-labeled occluded objects in the validation dataset based on a randomly designated ego agent.
From a bird's-eye view, and given the positions and sizes of all agents, we compute the occupancy grid following \cite{itkina2022multi}. 
Objects within the occluded region are labeled as \textit{occluded}, as demonstrated in Fig. \ref{fig:label_occlusion}.
The network is initially trained using the original training data, after which it is tested on this postprocessed validation dataset.
The training uses a bird's-eye view without occlusions, while the validation set includes realistic real-world occlusions as seen from the vehicle's perspective.

\begin{figure}
  \centering
  \includegraphics[width=1.0\linewidth]{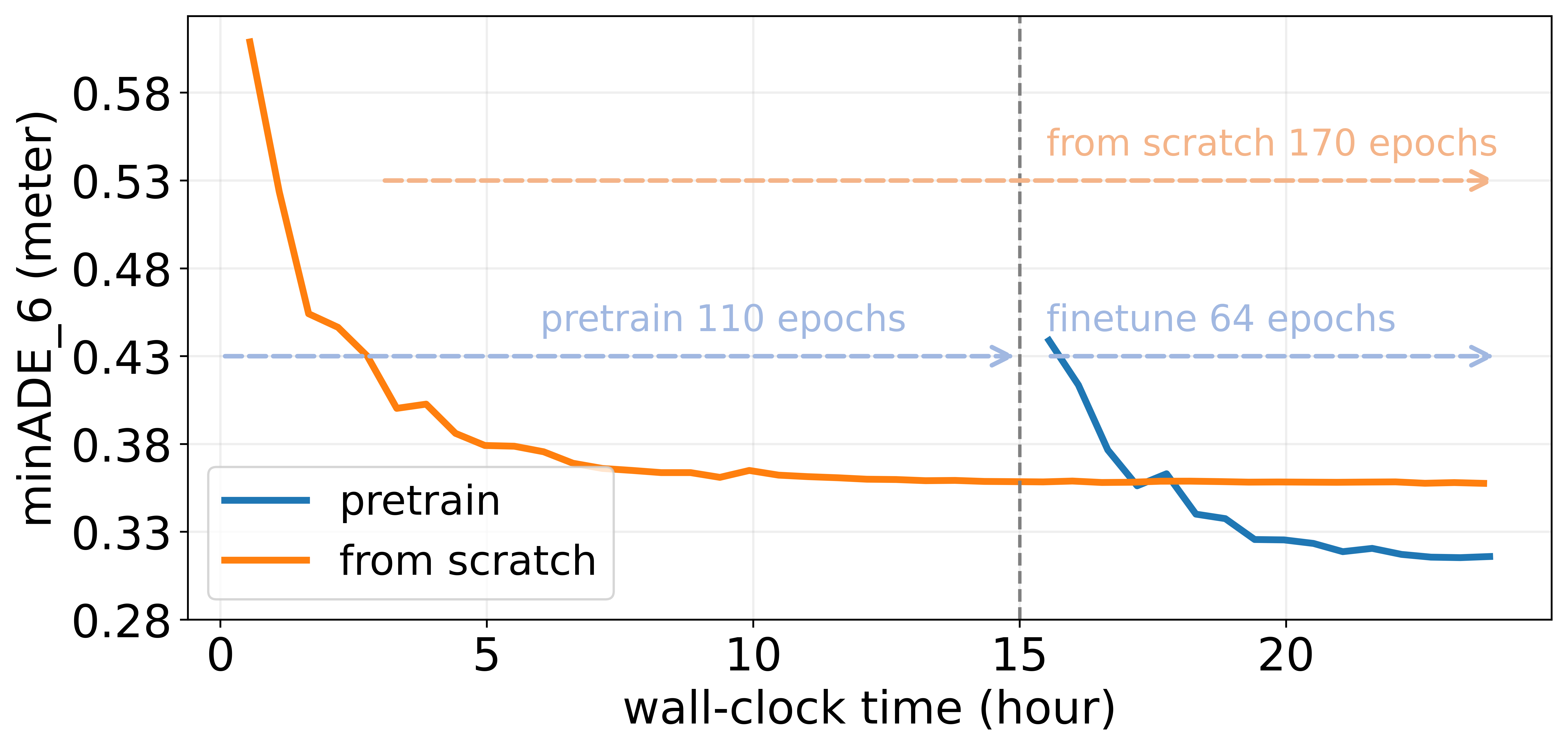}
  \caption{Conditional motion prediction on Argoverse dataset. Pretraining with fine-tuning is more accurate than training from scratch. The model is Autobot-joint. X-axis represents relative wall training time (4$\times$A100 GPUs), and Y-axis represents the $minADE_6$. The pretraining is done with 75\% pointwise masking.}
  \label{fig:time-result}
\end{figure}

\subsection{Masking Strategy}
We have conducted extensive testing to assess the impact of different masking strategies on performance. The results of these ablation experiments are presented in Table \ref{tab:ablation_pointwise_mask}. Table \ref{tab:mask_ratio} displays the outcomes of tests utilizing varying mask ratios and profiles for the pretraining task. Interestingly, for pointwise masking, ratios of 50\% and 75\% yielded superior results. Conversely, for both patchwise and time-only masks, a 25\% ratio demonstrated the best performance. Among the tested profiles, point masking proved most effective.
In regards to frozen encoder weights, the experiment shows that the unfrozen encoder achieves better results (Table \ref{tab:frozen_encoder}). We also test with different encoder sizes. The default Autobot model utilizes 2 sets of axial attention for temporal and social relations ($\sim$1,160,320 parameters for the encoder). Despite extending the size to include 4 and 6 sets, larger networks did not result in improved performance as demonstrated in Table \ref{tab:encoder_size}. This could be attributed to the Argoverse1 dataset size which is not large.

To ensure fair comparison between pretraining and training from scratch, we perform experiments over comparable time periods and on identical devices. 
As an example, the conditional motion prediction results for Argoverse dataset (Fig. \ref{fig:time-result}) show that pretraining 
achieves better results and converges faster. 
Our experiments also show that it can learn other tasks from that same pretrained network at a faster rate and to better results.

\begin{table}[t]
\centering
\caption{Performance comparison of different models on nuScenes dataset. Here we use the baseline results of Autobot without ensemble to maintain a fair comparison.}
\label{tab:motion_pred_nus}
\adjustbox{width=\linewidth}{
\begin{tblr}{
column{2-4} = {l}, 
hline{1-2,6,8} = {-}{},
}
Method         & minADE\_5 ↓ & minADE\_10 ↓ & \begin{tabular}[c]{@{}c@{}}Miss Rate↓\\(Top 5)\end{tabular}  \\
GOHOME~\cite{gilles2022gohome}         & 1.42        & 1.15         & 0.57        \\
THOMAS~\cite{gilles2021thomas}         & 1.33        & 1.04         & 0.55        \\
PGP~\cite{deo2021multimodal}& 1.27        & 0.94         & 0.52        \\
FRM~\cite{park2023leveraging}& 1.18        & 0.88         & 0.48        \\
\begin{tabular}[c]{@{}l@{}}Autobot~\cite{girgis2022latent-autobot}\\(Baseline, w/o ensemble)\end{tabular}
& 1.43        & 1.05         & 0.66        \\
\begin{tabular}[c]{@{}l@{}}Autobot w/ [P]\\(Ours)\end{tabular}
& 1.38  (\textcolor[rgb]{0,0.439,0.753}{3.5\%})        & 0.98 (\textcolor[rgb]{0,0.439,0.753}{6.7\%})       & 	0.60  (\textcolor[rgb]{0,0.439,0.753}{9.1\%})        
\end{tblr}
}
\end{table}

\begin{table}[t]
\centering
\caption{Prediction performance of different models on Argoverse validation dataset. Note that the results presented have been obtained from our own training runs.}
\label{tab:motion_pred_argo}
\begin{tblr}{
  hline{1-2,4,6} = {-}{},
}
Method         & minADE\_6 ↓                                   & minFDE\_6 ↓                                   \\
Autobot (Baseline)       & 0.722                                         & 1.288                                         \\
Autobot w/ [P] (Ours) & 0.694 (\textcolor[rgb]{0,0.439,0.753}{3.9\%}) & 1.229 (\textcolor[rgb]{0,0.439,0.753}{4.6\%}) \\
HivT (Baseline)           & 0.760                                         & 1.215                                         \\
HivT w/ [P] (Ours)   & 0.723 (\textcolor[rgb]{0,0.439,0.753}{4.9\%}) & 1.196 (\textcolor[rgb]{0,0.439,0.753}{1.6\%}) 
\end{tblr}
\end{table}

\begin{table}[t]
    \centering
    \caption{Conditional motion prediction results for Argoverse and nuScenes validation data.}
    \begin{tblr}{
      row{1} = {c},
      cell{1}{1} = {r=2}{},
      cell{1}{2} = {c=2}{},
      hline{1,3,5} = {-}{},
      hline{2} = {2-3}{},
    }
    Method             & Argoverse                                      &                                                \\
                   & minADE\_6 ↓                                    & minFDE\_6 ↓                                    \\
    Autobot (Baseline) & 0.358                                          & 1.016                                          \\
    Autobot w/ [P] (Ours) & 0.315 (\textcolor[rgb]{0,0.439,0.753}{12.0\%}) & 0.912 (\textcolor[rgb]{0,0.439,0.753}{10.2\%}) \\
    \\[3pt] 
    \end{tblr}
    \begin{tblr}{
      row{1} = {c},
      cell{1}{1} = {r=2}{},
      cell{1}{2} = {c=2}{},
      hline{1,3,5} = {-}{},
      hline{2} = {2-3}{},
    }
    Method             & nuScenes                                      &                                               \\
                   &    minADE\_10 ↓                                & minFDE\_10 ↓                                 \\
    Autobot (Baseline)       & 1.436                                           &  3.105                                      \\
    Autobot w/ [P] (Ours) &  1.309 (\textcolor[rgb]{0,0.439,0.753}{8.8\%}) & 2.953 (\textcolor[rgb]{0,0.439,0.753}{4.9\%})
    \end{tblr}

    \label{tab:conditional_pred_result}
\end{table}

\renewcommand{\arraystretch}{1.3}
\begin{table}[t]
\centering
\caption{Prediction results for the postprocessing INTERACTION validation dataset, focusing on scenarios with occlusion.}
\label{tab:occlusion}
\begin{tblr}{
  hline{1-2,4} = {-}{},
}
               & minADE\_6 ↓                                    & minFDE\_6 ↓                                    \\
Autobot (Baseline)        & 3.034                                          & 5.693                                          \\
Autobot w/ [P] (Ours) & 1.862 (\textcolor[rgb]{0,0.439,0.753}{38.6\%}) & 3.678 (\textcolor[rgb]{0,0.439,0.753}{35.4\%})
\end{tblr}
\end{table}
\renewcommand{\arraystretch}{1.0}

\subsection{Motion Prediction}
We have integrated our framework into the nuScenes (Table~\ref{tab:motion_pred_nus}) and Argoverse (Table~\ref{tab:motion_pred_argo}) datasets for motion prediction. 
The results indicate that the implementation of random masking pretraining enhances performance.
In nuScenes, our approach achieves comparable results to other state-of-the-art methods. Compared to the baseline, the application of random masking showed marked improvements in the metrics including $minADE_5$, $minADE_{10}$ and miss rate for the Top 5 in 2 meters, with percentage decreases of 3.5\% and 6.7\% and 9.1\% respectively. Note that in order to maintain a fair comparison, the Autobot baseline we utilize does not include ensemble operations, as these are not used in our post-processing steps.
In Argoverse, we incorporate two methods- Autobot-joint and HivT. Both of them show a positive impact of masked pertaining, resulting in an decrease of $minADE_6$, $minFDE_{6}$ by 3.9\% and 4.6\% for Autobot, and 4.9\% and 1.6\% for HiVT. Note that for HiVT, we prioritized speed and trained on four GPUs, resulting in lower of performance than training on a single GPU. However, our comparison is conducted under the same environment and settings.

\subsection{Conditional Motion Prediction}
We evaluate conditional motion prediction on nuScenes and Argoverse datasets with Autobot again.
Given the history information and the ego vehicle's desired future trajectories, the task is to predict all other agents' possible future trajectories.
The results for this task are shown in Table. \ref{tab:conditional_pred_result}. For Argoverse, it reduces the minADE6 and minFDE6 by 12.0\% and 10.2\%, respectively. For nuScenes, it reduces minADE10 and minFDE10 by 8.8\% and 4.9\%. 
Given that the Argoverse data features higher frequency and more waypoints, it is plausible that random masking exhibits superior performance as the input size expands.

\subsection{Occlusion Handling}
We use the postprocessed validation INTERACTION dataset to evaluate the efficacy of Autobot in complex scenarios as well as the benefits of random masking.
The network is trained with regular INTERACTION training data. However, during the inference time, the network can only access the agent's waypoints annotated as visible. Thus, for the agents that are partially occluded (Section \ref{sec:exp-dataset}), the network can only see incomplete history. We then measure how the network can capture such partially occluded agents' future trajectories.
As shown in Table \ref{tab:occlusion}, the use of random masking enhances the network's capability to predict the partially occluded agent's future trajectory, with improvements exceeding 30\% for both $ADE$ and $FDE$.

The results are not surprising as the pretraining is a sort of random synthetic occlusion (as opposed to the actual realistic occlusions that we model in the validation set).  Therefore, the pretrained network has a considerable advantage over a network simply trained with bird's eye view data and no occlusions.

\section{Conclusion}
In this paper, we propose a simple and effective random mask pretraining framework which facilitates the motion prediction task in general and conditional motion prediction. Furthermore, our framework largely improves the prediction accuracy for occlusion scenarios.
The self-supervised learning and masked autoencoder can be explored further with state-of-the-art techniques in the field of motion prediction for autonomous driving.
Additionally, exploring new auxiliary tasks within the self-supervised learning domain offers exciting possibilities for further advancements.
We think that exploring self-supervised learning may be beneficial as the volume of motion prediction data expands.

\section*{Acknowledgement}

This work\footnote{We have used ChatGPT for editing and polishing author-written text.} was funded by Vinnova, Sweden (research grant). The computations were enabled by the supercomputing resource Berzelius provided by National Supercomputer Centre at Linköping University and the Knut and Alice Wallenberg foundation, Sweden.

\bibliographystyle{IEEEtran}
\bibliography{IEEEabrv,ref}

\end{document}